\documentclass[letterpaper]{article} 
\usepackage[submission]{aaai24}  
\usepackage{times}  
\usepackage{helvet}  
\usepackage{courier}  
\usepackage[hyphens]{url}  
\usepackage{graphicx} 
\usepackage{natbib}  
\usepackage{caption} 
\usepackage{algorithm}
\usepackage{algorithmic}
\usepackage{amssymb}
\usepackage{multirow} 
\usepackage{amsmath}

\usepackage{newfloat}
\usepackage{listings}
\DeclareCaptionStyle{ruled}{labelfont=normalfont,labelsep=colon,strut=off} 
\floatstyle{ruled}
\newfloat{listing}{tb}{lst}{}
\floatname{listing}{Listing}
\title{Overlap Bias Matching is Necessary for Point Cloud Registration}
\author{
    Written by AAAI Press Staff\textsuperscript{\rm 1}\thanks{With help from the AAAI Publications Committee.}\\
    AAAI Style Contributions by Pater Patel Schneider,
    Sunil Issar,\\
    J. Scott Penberthy,
    George Ferguson,
    Hans Guesgen,
    Francisco Cruz\equalcontrib,
    Marc Pujol-Gonzalez\equalcontrib
}
\affiliations{
    \textsuperscript{\rm 1}Association for the Advancement of Artificial Intelligence\\

    1900 Embarcadero Road, Suite 101\\
    Palo Alto, California 94303-3310 USA\\
    proceedings-questions@aaai.org
}

\title{Overlap Bias Matching is Necessary for Point Cloud Registration}
\author {
    First Author Name\textsuperscript{\rm },
    Second Author Name\textsuperscript{\rm 2},
    Third Author Name\textsuperscript{\rm 1}
}

\usepackage{bibentry}

\begin{document}

\maketitle

\begin{abstract}
Point cloud registration is a fundamental problem in many domains. Practically, the overlap between point clouds to be registered may be relatively small. Most unsupervised methods lack effective initial evaluation of overlap, leading to suboptimal registration accuracy. To address this issue, we propose an unsupervised network Overlap Bias Matching Network (OBMNet) for partial point cloud registration. Specifically, we propose a plug-and-play Overlap Bias Matching Module (OBMM) comprising two integral components, overlap sampling module and bias prediction module. These two components are utilized to capture the distribution of overlapping regions and predict bias coefficients of point cloud common structures, respectively. Then, we integrate OBMM with the neighbor map matching module to robustly identify correspondences by precisely merging matching scores of points within the neighborhood, which addresses the ambiguities in single-point features. OBMNet can maintain efficacy even in pair-wise registration scenarios with low overlap ratios. Experimental results on extensive datasets demonstrate that our approach's performance achieves a significant improvement compared to the state-of-the-art registration approach.
\end{abstract}

\section{Introduction}

Point cloud registration is a crucial task in the fields of robotics and computer vision. Its primary objective is to align multiple point clouds through appropriate rigid transformations~\cite{agarwal2011building,schonberger2016structure}. This process finds widespread applications in areas such as 3D map reconstruction, object detection, and autonomous driving~\cite{wong2017segicp,deschaud2018imls,zhang2014loam}. However, achieving accurate registration becomes challenging due to factors like occlusions, different viewpoints, and inaccurate scanning~\cite{deng20193d,qi2017pointnet++}.

In recent years, with the discriminative representation capabilities of deep learning, deep point cloud registration methods have garnered increasing research attention~\cite{tombari2010unique,yang2017foldingnet}.  These methods focus on supervised learning of rigid transformations, requiring a large number of real ground truth transformations as supervised signals for model training ~\cite{pais20203dregnet,fu2021robust, ali2021rpsrnet,wang2020unsupervised}.  As a result, the training costs escalate, hindering their practical application in real-world scenarios.  To address these challenges, various approaches have been proposed.   For instance, STORM~\cite{9705149} proposed a structure-based overlap matching method for local point cloud registration, detecting overlapping points using structural information and generating precise partial correspondences based on feature similarity. However, the network structure of STORM is complex and requires effective supervision, leading to poor generalization in real point cloud scenes.   RIENet~\cite{2022Reliable} proposed an unsupervised point cloud registration network with reliable internal evaluation, consisting of a matching refinement module and an internal evaluation module.   However, it fails to handle partial-to-partial point cloud registration effectively.

\begin{figure}
	\centering
	\includegraphics[width=0.95\columnwidth]{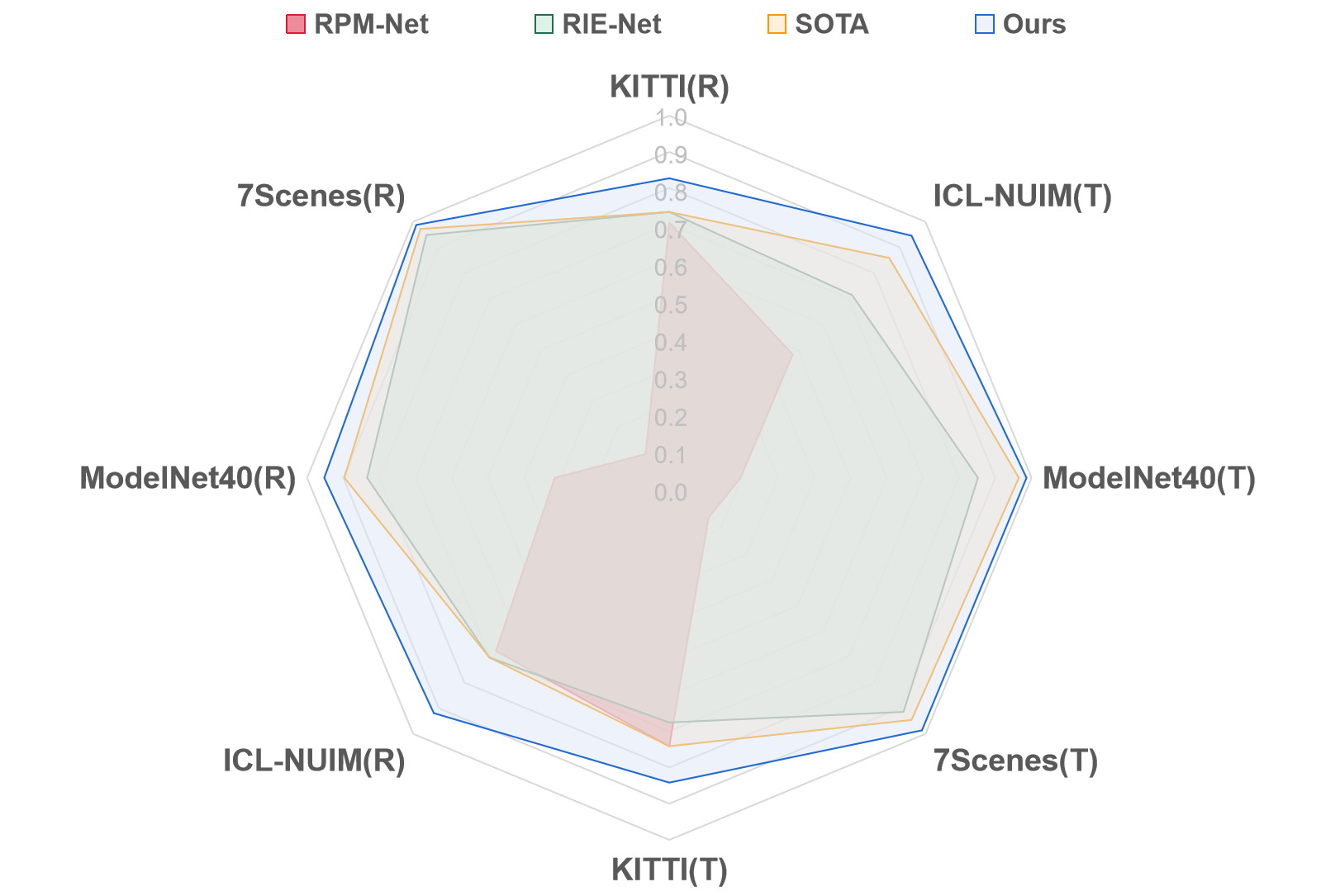} 
	\setlength{\belowcaptionskip}{-0.5cm}
	\caption{Our method exhibits consistently superior performance across all four 
        datasets through the utilization of min-max normalization.
	}
	\label{fig1}
	\vspace*{-0.6\baselineskip}
\end{figure}

In our paper, we present an Overlap Bias Matching Module (OBMM) suitable for most point cloud registration tasks and propose an unsupervised framework for partial point cloud registration based on this module.  Specifically, we use the output of the feature extraction module as the input to OBMM.  After passing through a differentiable overlap sampling module, we detect common points in the point cloud using point-wise features to generate partial correspondences. We employ the point cloud structure representing the overlap region along with point-wise features for bias prediction, obtaining matching coefficients for the overlapping parts of the two point clouds. These coefficients are used in the weighted SVD for rigid transformation estimation. By fusing global features and locally sampled estimates, we construct a neighbor map matching module to  merge the discriminating matching scores of adjacent points to clarify ambiguous single-point features and achieve reliable correspondence recognition. The enriched multi-neighborhood information effectively promotes the construction of the matching graph and provides high-quality correspondences.The experimental results are presented as shown in Figure~\ref{fig1} for demonstration. Our contributions can be summarized as follows:

\begin{itemize}
\item We present a plug-and-play Overlap Bias Matching Module (OBMM) consisting of an overlap sampling module and a bias prediction module. This module can effectively analyze the overlapping regions of point clouds in a structured manner and compute bias coefficients.
\item We introduce a neighbor map matching module that addresses the ambiguity in single-point matching. An unsupervised point cloud registration network is proposed for partial point clouds by integrating with OBMM.
\item Extensive experimental results across a variety of benchmark datasets consistently demonstrate that our method attains state-of-the-art performance in registration.
\end{itemize}

\section{Related Work}
\subsection{Point Cloud Registration}
Recently, deep learning has been successfully applied to the point cloud registration which can get the rigid transformation in an end-to-end manner~\cite{fischer2021stickypillars,sarode2019pcrnet,gojcic2020learning,choy2020deep,bai2021pointdsc}. There are some supervised methods for point cloud registration through good initial judgment~\cite{lu2019deepvcp,yuan2020deepgmr,sarode2020masknet,huang2021predator}. Following the soft matching map based methods ~\cite{wang2019deep,wang2019prnet,mellado2014super}, RPM-Net ~\cite{yew2020rpm} uses the Sinkhorn layer to enforce the doubly stochastic constraints on the matching map for reliable correspondences~\cite{li2020iterative}. proposes an iterative distance-aware similarity matrix convolution network with two-stage point elimination technique for point cloud registration.Additionally, there are some end-to-end unsupervised point cloud registration methods ~\cite{kadam2020unsupervised,feng2021recurrent,li2019pc,el2021unsupervisedr,huang2020feature,jiang2021CEMNet}. ~\cite{yang2020mapping,groueix2019unsupervised,fischler1981random} employ cycle consistency across the pairwise point clouds for points matching, which cannot be trained directly on the partial data. RIENet~\cite{2022Reliable}introduces an internally assessed unsupervised method for point cloud registration. However, its performance is suboptimal for aligning partial point clouds with low overlap.

\subsection{Neighborhood Consensus.} Recently, efforts on neighborhood consensus have been made to establish correspondences between images ~\cite{aoki2019pointnetlk,qi2017pointnet, li2021pointnetlk, kadam2020unsupervised}. Those methods first build a 4D correlation tensor, then employ the 4D CNN on this 4D tensor to achieve the neighborhood consensus~\cite{yang2013go,chen1999ransac,le2019sdrsac}. Since the 4D tensor are inherently contains neighborhood information, the convolution filters can capture patterns in the pairwise matches of two neighborhoods. However, the operations used in images cannot be directly applied to disordered point clouds. Based on neighborhood consensus, we develop two modules to obtain accurate correspondences and distinguish outliers for robust registration~\cite{phillips2007outlier,bouaziz2013sparse,segal2009generalized}.
	Since point clouds are disordered and irregular, it cannot directly achieve neighborhood consensus on point clouds with the operations used in images, so we design two modules suitable for point cloud with neighborhood consensus~\cite{rusu2008aligning, rusu2009fast, tombari2010unique}.
	 Among them~\cite{yang2020mapping}, with the sparse convolutions and two-stage correspondence relocalisation achieves less memory consumption and more accurate localisation. 
\begin{figure*}
	\centering
	\includegraphics[width=0.96\textwidth]{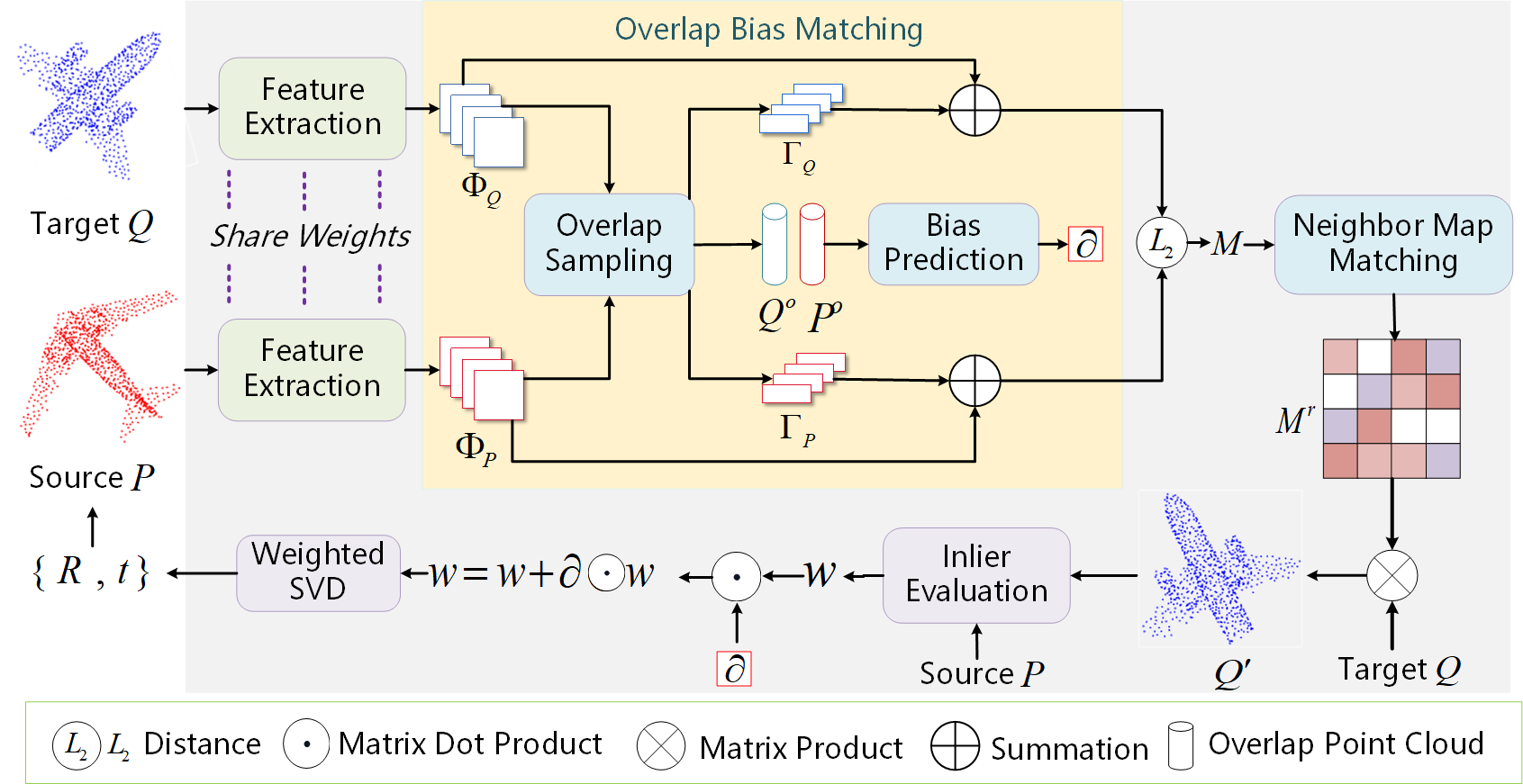} 
	\caption{The overall pipeline of OBMNet.  Given the input point clouds, It first extracts the point-wise features.  Through overlapping sampling, we determine the overlap between the source point cloud \textbf{P} and the target point cloud \textbf{Q}.  Then, we get a bias coefficient $\alpha$ by bias prediction, which is applied in the subsequent SVD computation.  The subsequent neighbor matching module refines the point-wise feature representation, yielding $\boldsymbol{M_r}$, which is utilized for generating the pseudo-target point cloud and determining rigid transformations \{R,t\} in conjunction with $\alpha$.  }
	\label{fig2}
	\vspace{-.3cm}
\end{figure*}

\section{Method}
Given a source point cloud $\mathbf{P}=\{\mathbf{p}_i\in\mathbb{R}^3\mid i=1,...,N\}$ and a target point cloud $\mathbf{Q}=\left\{\mathbf{q}_j\in\mathbb{R}^3 \mid j=1,...,M\right\}$.  For partial point cloud registration, our proposed method performs feature extraction to generate per-point features ${\Phi}_{\mathbf{P}}$ and ${\Phi}_{\mathbf{Q}}$ as Figure 2 shows. Global features are constructed as overlapping representations $\boldsymbol{{T}_P}$ and $\boldsymbol{{T}_Q}$  by OBMM, while predicting coefficients $\boldsymbol{\alpha}$ for the bias distribution of the two frames of overlapping point clouds.  By fusing the overlapping representations of point-wise features and global per-point features, we build a matching map $\boldsymbol{M}$. Using the neighbor map matching module, we calculate the relationship coefficients between the target domain and neighboring points, and weight the discriminative matching scores of neighboring points to merge them into the matching graph, thereby obtaining a more reliable corresponding matching map $\boldsymbol{M_r}$. With this, we predict pseudo-target point cloud $\mathbf{Q}'$, and through an internal evaluation module, obtain confidence scores $\mathbf{w}_{i}$ for each pseudo-matched pair in the target point cloud.  Finally, we incorporate the bias coefficient $\alpha$ as an input to the SVD process for solving the rigid transformation \{R, t\}. This iterative training process continues until a predefined threshold is met.

\subsection{Overlap Bias Matching Module}
This paper introduces a novel approach utilizing a differential sampling mechanism within an overlap bias matching module for partial point cloud registration.    The proposed method comprises an overlap sampling network based on Gumbel-Softmax and a bias prediction sub-network that estimates the current overlap parameters.

\subsubsection{Overlap samping.}
Given the source point cloud \textbf{P} and the target point cloud \textbf{Q}, in order to enhance the discriminative nature of the learned point-wise features, we employ multiple densely connected EdgeConv layers for feature extraction, yielding per-point representations $\Phi_{\mathcal{P}} \in \mathbb{R}^{N \times D}$ and $\Phi_{\mathcal{Q}} \in \mathbb{R}^{M \times D}$, where $\emph{N, M}$ represents the number of points and $\emph{D}$ represents the embedding dimension. These representations are then utilized as inputs to the overlap sampling module, which serves to identify the overlapping regions between point clouds \textbf{P} and \textbf{Q}. The details of our overlap prediction module are illustrated in Figure 3.

The overlap sampling module relies on a differentiable sampling mechanism. By integrating the feature representations of two frames of point clouds and constructing fully connected layers with adapted channel sizes, we obtain class probabilities $\boldsymbol{\pi}=\left\{\pi_{1}, \cdots \pi_{m}\right\}$ . Given the Gumbel noise samples $g_{1} \ldots g_{m}$ drawn from Gumbel(0, 1) distribution and the class probabilities, we can compute a categorical sample $\boldsymbol{z}$ as follows:

\begin{figure*}
	\centering
	\includegraphics[width=0.90\textwidth]{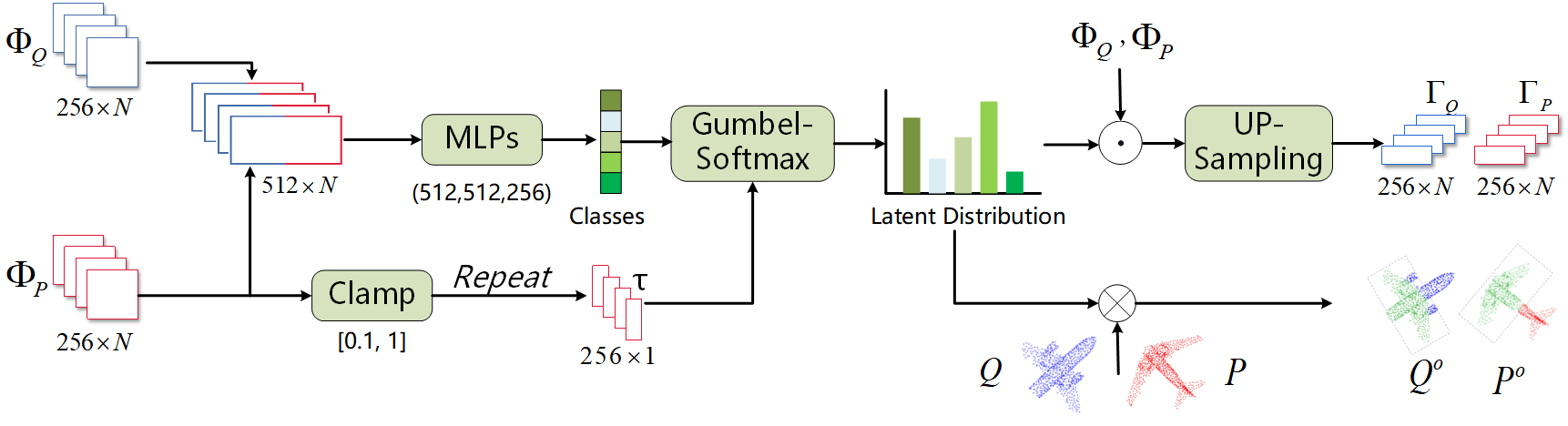} 
	\caption{Given the feature inputs ${\Phi}_{\mathbf{Q}}$ and ${\Phi}_{\mathbf{P}}$, we first blend the features from two frames, which is achieved by constructing fully connected layers with channel dimensions adapted accordingly. This process yields class probabilities along with a temperature parameter $\tau$, which are then employed as inputs for Gumbel-Softmax to derive a distribution describing the latent shared structure within the input point clouds. Finally, we obtain a representation of the overlapping region through sampling.}
	\label{fig3}
	\vspace{-.3cm}
\end{figure*}

\vspace{-.5cm}
\begin{equation}
	z  =  one\_hot \left(\underset{j}{\arg \max }\left[\left(g_{j}+\log \pi_{j}\right) / \tau\right]\right),
\end{equation}

\noindent where the parameter $\boldsymbol{\tau}$ controls the sharpness of the categorical distribution. 

We compress the feature representation ${\Phi}_{\mathbf{P}}$ of the source point cloud, aligning its dimensions with the input class probabilities, with tensor size tailored to the structure of the original point cloud. Furthermore, in the network architecture, we replace the arg-max operation with a soft-max function approximation, rendering Gumbel-Softmax differentiable. This approach ensures the continuity of the optimization process. Then a distribution indicative of the latent overlapping structure within the input point cloud can be generated, leading to the visualization of overlapping samples. The sampled points ${\mathcal{Q}}^{o}$, ${\mathcal{P}}^{o}$ and their corresponding point-wise features $\Gamma_{Q}, \Gamma_{P}\in \mathbb{R}^{K \times D}$ yield the following outcomes:

\vspace{-.5cm}
\begin{equation}
	\{\mathcal{Q}^{o}, \mathcal{P}^{o}\}  =\text { gumbel\_softmax }(\boldsymbol{\pi}, \tau) \cdot \{\boldsymbol{Q},\boldsymbol{P}\}, \\
\end{equation}

\vspace{-.5cm}
\begin{equation}
	\{\Gamma_{Q},\Gamma_{p}\} =Up(\text { gumbel\_softmax }(\boldsymbol{\pi}, \tau) \cdot \{\Phi_{Q}, \Phi_{P}\}),
\end{equation}
where Up denotes Up-Sampling, and K represents the number of points sampled from the distribution. It's important to note that each sample is independent. These K samples represent the overlap between P and Q.

\begin{figure}
	\centering
	\includegraphics[width=0.46\textwidth]{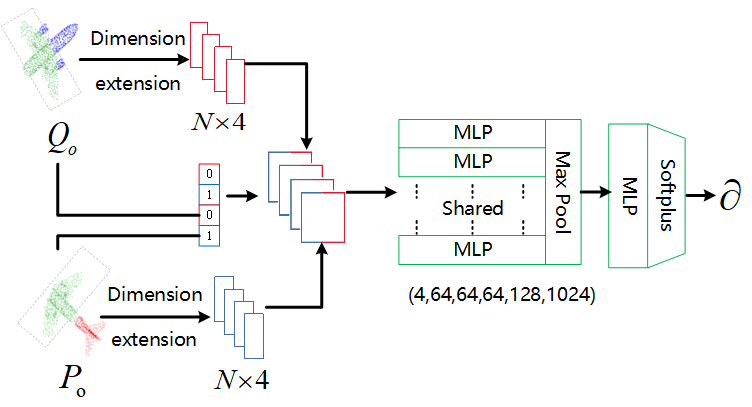} 
	\caption{Given the input overlapping point clouds ${\mathcal{Q}}^{o}$ and ${\mathcal{P}}^{o}$ , we concatenate the feature tensors through dimension extension. This involves augmenting the fourth dimension with columns containing 0 or 1. Afterward, a fully connected layer is utilized to produce the overlap coefficient $\alpha$ for the two frames of point clouds.}
	\label{fig4}
	\vspace{-.3cm}
\end{figure}

\subsubsection{Bias prediction.}
In previous point cloud registration algorithms, most SVD computations primarily focused on the pseudo-correspondence relationship, while overlooking the bias relationship between two point clouds.   To address this, we introduce a secondary neural network that takes two point clouds as inputs and predicts the bias coefficients for the current iteration.   These coefficients are then used as weights for the overlapping correspondences during SVD computation.   Specifically, we concatenate the two point clouds to form a (2N, 4) matrix, where the fourth column indicates the origin of each point (from either of the two point clouds) using an extended column containing 0 and 1.   This constructed matrix tensor is fed into a series of fully connected layers comprising the point network.   To ensure that the predicted $\alpha$ coefficients are positive, we employ a softplus activation for the final layer as Figure 4 shows.

\subsection{Neighbor Map Matching Module}
Point cloud registration methods based on feature matching often utilize individual point feature distances as scores for constructing matching maps. However, non-corresponding points with similar point features can mislead these methods into making erroneous correspondence estimations. To mitigate this issue, the neighbor map matching module addresses the problem by aggregating discriminative matching scores of neighboring points, thereby enhancing the clarity of ambiguous single-point features and enabling robust correspondence identification.For each point cloud pair $\mathbf{p}_i$ and $\mathbf{q}_j$, we calculate their point-wise matching score with the normalized negative feature distance by calculating the index ${src\_idx}$ of {k} nearest neighbor points for each point in the input point cloud \textbf{P} as below:
 
\begin{figure}
	\centering
	\includegraphics[width=0.46\textwidth]{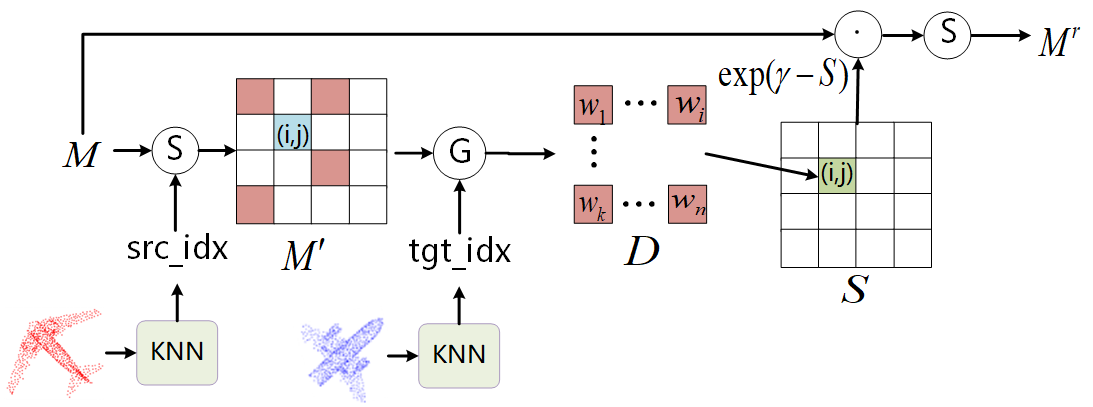}  
	\caption{By utilizing the indices of the source point cloud and the target point cloud, combined with a matching map $\emph{M}$. The result $M^r$ is achieved by weightedly merging the matching scores of neighboring points to refine ambiguous singular point features.}
	\label{fig5}
	\vspace{-.5cm}
\end{figure}

\begin{equation}\label{M1}
	\mathbf{M'}_{i,j}= \operatorname{softmax} \left(\left[-\mathbf{M}_{i,1},\ldots,-\mathbf{M}_{i,M}\right]\right)_{j}[{src\_idx}, : ],
\end{equation}
where $\mathbf{M}_{i,j} = \left\|{\Phi}_{\mathbf{p}_i} - {\Phi}_{\mathbf{q}_j}\right\|_2$ denotes the euclidean distance between the single-point features of points $\mathbf{p}_i$ and $\mathbf{q}_j$. 

Given the matching matrix $\mathbf{M}_{i,j}^{'}$ as input, we compute the indices ${tgt\_idx}$ of the {k} nearest neighbor points for each point in the input point cloud \textbf{Q}. Using these index tensors, we gather features and reshape the collected features. We then calculate pairwise distances between these reshaped features, resulting in a weighted distance measure \textbf{D}, which is defined as follows:

\vspace{-.5cm}
\begin{equation}\label{eq0}
	\begin{split}
	d_{i, j}=&\sum_{p_{i} \in P, j=1}^{M} \| \text { gather }\left(M_{i, j}^{'}, \operatorname{tg} t_{-} i d x_{i}\right) \|_{2}^{2},  \\
	D_{i, j}=& ({1 / d_{i, j}+\beta}) / \left({\sum_{j=1}^{M} (1 / d_{i, j}+\beta )}\right),
	\end{split}
\end{equation}
where $d_{i, j}$ represents pairwise distances between the collected features. The symbol $\beta$ corresponds to a very small value, typically set to $e^{-6}$ in practical applications.Based on the estimated point-level matching scores, their neighborhood-level scores can be computed by weighting the corresponding scores of the surrounding points:

\begin{equation}\label{eq1}
	\mathbf{S}_{i,j}= \frac{D_{i, j}}{K}\sum_{\mathbf{p}_{i} \in \mathcal{N}_{\mathbf{p}_i}}\sum_{\mathbf{q}_{j} \in \mathcal{N}_{\mathbf{q}_j}}\mathbf{M'}_{i,j},
\end{equation}
where $\mathcal{N}_{\mathbf{p}_i}$ denotes the $k$-nearest neighbor points around the point $\mathbf{p}_i$, $\mathcal{N}_{\mathbf{q}_j}$ for the same reason. The high neighborhood score means the surrounding points consistently tend to have large matching probabilities and vice versa. Based on the fact that the real correspondence usually owns the high neighborhood similarity while this similarity for incorrect correspondence is prone to be low,we can effectively exploit this neighborhood score to identify the non-corresponding point pair.
Finally, the neighbor matching map can be formulated as:

\vspace{-.2cm}
\begin{equation}\label{eq0}
	\begin{split}
	\mathbf{M}_{i,j}^{r}&= \operatorname{softmax} \left(\left[-\mathbf{M}^{e}_{i,1},\ldots, -\mathbf{M}^{e}_{i,M}\right]\right)_{j},\\
	\mathbf{M}_{i,j}^{e} &= \exp \left(\mathbf{\gamma} - \mathbf{S}_{i,j}\right) * \mathbf{M}_{i,j},
	\end{split}
\end{equation}
\vspace{.1cm}
where the refined feature distance $\mathbf{M}_{i,j}^{e}$ is negatively related to the neighborhood score  $\mathbf{S}_{i,j}$.and we design an exponential function to control the changing ratio. The hyper-parameter $\gamma$ controls the influence of the neighborhood consensus. The process is shown in Figure 5.

\subsection{Loss Function}
In each iteration, we can obtain a set of pseudo correspondences and their corresponding inlier confidence $\mathbf{w}$. Then, we employ the weighted SVD on the pseudo correspondences with $\mathbf{w}$ and $\mathbf{\alpha}$ to estimate the rigid transformation $\left\{\mathbf{R},\mathbf{t}\right\}$. Under the unsupervised setting, we construct the following unsupervised loss functions for model optimization.
	
	\noindent \textbf{Global alignment loss.} 
	We first exploit the alignment error based loss function to train our model.  To handle the partially-overlapping problem well, we integrate the Huber function, a robust loss function insensitive to the outliers. The Huber function based overall alignment loss can be formulated as:

	\begin{equation}
	\begin{split}
	\mathcal{L}_{g}=&\sum_{\mathbf{p}'_{i} \in \mathbf{P}^{\prime}} \vartheta \left(\min _{\mathbf{q_i} \in \mathbf{Q}}\|\mathbf{p_i}^{\prime}-\mathbf{q_i}\|_{2}^{2}\right) \\
	+&\sum_{\mathbf{q_i} \in \mathbf{Q}} \vartheta \left(\min _{\mathbf{p_i}' \in \mathbf{P}^{\prime}}\|\mathbf{q_i}-\mathbf{p_i}'\|_{2}^{2}\right),
	\end{split}
	\end{equation}

\noindent where $\mathbf{P}'$ denotes the transformed source point cloud with the estimated transformation and  $\vartheta$  is the huber function. 
	
	\noindent \textbf{Neighborhood agreement loss.} 
To prevent the model from becoming trapped in local optima, we introduce the robust KNN-based neighborhood consensus loss along with the spatial consistency loss, aimed at further mitigating the impact of outliers. By selecting the $\emph{k}$ pairs of points with the highest weights determined through overlap prediction, we acquire overlapping point clouds: $\mathbf{X}\in \mathbb{R}^{k\times3}$ and $\mathbf{Y}\in \mathbb{R}^{k\times3}$. This enables us to formulate the neighborhood agreement loss using the transformation \{R, t\}:

	\begin{equation}
	\begin{split}
	 \mathcal{L}_{n} = \sum_{\mathbf{x}_i \in \mathbf{X}, \mathbf{y}_i \in \mathbf{Y}} \sum_{\mathbf{p}_j \in \mathcal{N}_{\mathbf{x}_i}, \mathbf{q}_j \in \mathcal{N}_{\mathbf{y}_i} } \|\mathbf{R}\mathbf{p}_j + \mathbf{t} - \mathbf{q}_j \|_{2}.	
	\end{split}
	\end{equation}
	
	\noindent \textbf{Spatial consistency loss.} 
	In order to further eliminate the spatial gap between the pseudo target correspondence and the real target correspondence for each selected $\mathbf{x}_i \in \mathbf{X}$, we exploit the cross-entropy based spatial consistency loss to sharpen their matching distributions as below:
	\begin{equation}\label{eq}
	\mathcal{L}_s = -\frac{1}{|\mathbf{X}|}\sum_{\mathbf{x}_i\in\mathbf{X}}\sum_{j=1}^M \mathbb{I}\left\{j=\arg\max_{j'}\mathbf{M}^{r}_{i,j'}\right\}\log\mathbf{M}^{r}_{i,j},
	\end{equation}
	where $\mathbb{I}\left\{\cdot\right\}$ denotes the indicator function and we use the target point $\mathbf{q}_{j}$ with the largest matching probability to estimate the real target correspondence. By improving the matching probability of the ``real'' target correspondence ($\mathbf{M}^{r}_{i,j}\rightarrow 1$), the resulting pseudo target correspondence  tends to further spatially approach to the ``real'' target correspondence $\mathbf{q}_{j}$ well.  
	Finally, we utilize a comprehensive loss function as below to optimize our model:
	\begin{equation}
	\mathcal{L} = \mathcal{L}_g +  \mathcal{L}_{n} +  \mathcal{L}_{s}.
	\end{equation}

\begin{table*}[h]
\centering
\renewcommand\arraystretch{1.5}
\setlength{\tabcolsep}{2mm}{
\resizebox{2\columnwidth}{!}{
\begin{tabular}{ccccc|cccc|cccc}
\hline
\multirow{2}{*}{Model} & \multicolumn{4}{c|}{Same}                                                                                                                               & \multicolumn{4}{c|}{Unseen}     
    & \multicolumn{4}{c}{Noise}   \\ \cline{2-13} 
                        & MAE(R)          & MAE(t) & MIE(R)          & MIE(t)  & MAE(R)          & MAE(t) & MIE(R)         & MIE(t)  & MAE(R)          & MAE(t) & MIE(R)         & MIE(t)\\ \hline
ICP                                         & 3.4339          & 0.0114                         & 6.7706          & 0.0227                                             & 3.6099          & 0.0116                         & 7.0556         & 0.0228    &4.6441 &0.0167 &9.2194 &0.0333                      \\
FGR                                       & 0.5972          & 0.0021                         & 1.1563          & 0.0041                                      & 0.4579          & 0.0016                         & 0.8442         & 0.0032      &1.0676 &0.0036 &2.0038 &0.0072                   \\
DCP                                       & 1.3405          & 0.0075                         & 3.4980          & 0.0070                                 & 1.7736          & 0.0093                         & 3.8657         & 0.0090          &2.7653  &0.0137   &5.7763   &0.0279               \\
IDAM                                          & 0.4243          & 0.0020                         & 0.8170          & 0.0040                                         & 0.4809          & 0.0028                         & 0.9157         & 0.0055        &2.3076 &0.0124 &4.5332 &0.0246                 \\
RPM-Net                                      & 0.0051          & 0.00005                         & 0.0201          & 0.00006                                   & 0.0064          & 0.00014                         & 0.0207         & 0.00014         &0.0075  &0.0000  &0.0221  &0.0001                \\
STORM                                      & 0.0037          & 0.00002                         & 0.0341          & 0.00005                                    & 0.0052          & 0.00004                         & 0.0415         & 0.00008         &0.0067    &0.00016    &0.0248    &\textbf{0.00014}                 \\
RIENet                                        & 0.0033          & 0.00004                         & 0.0210          & 0.00005                              & 0.0059          & 0.00004                         & 0.0228         & 0.00011       & 0.0069  &\textbf{0.00014} & 0.0230  &0.00016                 \\ \hline
Ours                              & \textbf{0.0013} & \textbf{0.00001}                & \textbf{0.0190} & \textbf{0.00003}                        & \textbf{0.0016} & \textbf{0.00001}                & \textbf{0.0192} & \textbf{0.00003}     &\textbf{0.0048}   &\textbf{0.00014}  &\textbf{0.0221 }  &0.00015            \\ \hline
\end{tabular}
}
}
\caption{The registration results of different methods on ModelNet40.}
\end{table*}

\begin{table*}[h]
\centering
\renewcommand\arraystretch{1.5}
\setlength{\tabcolsep}{2mm}{
\resizebox{2\columnwidth}{!}{
\begin{tabular}{ccccc|cccc|lclc}
\hline
\multirow{2}{*}{Model} & \multicolumn{4}{c|}{ICL-NUIM}                                                                       & \multicolumn{4}{c|}{7Scenes}                                                                         & \multicolumn{4}{c}{KITTI}                                                                                                 \\ \cline{2-13} 
                       & MAE(R)          & MAE(t) & MIE(R)          & MIE(t)& MAE(R)          & MAE(t) & MIE(R)           & MIE(t) & \multicolumn{1}{c}{MAE(R)} & MAE(t) & \multicolumn{1}{c}{MIE(R)} & MIE(t)\\ \hline
ICP                    & 2.4022          & 0.0699                         & 4.4832          & 0.1410                         & 6.0091          & 0.0130                         & 13.0484          & 0.0260                         & 4.7433                     & 0.9174                         & 11.9982                    & 2.5732                         \\
FGR                    & 2.2477          & 0.0808                         & 4.1850          & 0.1573                         & 0.0919          & 0.0004                         & 0.1705           & 0.0008                         & 1.6777                     & 0.0352                         & 4.0467                     & 0.0762                         \\
DCP                    & 5.7619          & 0.1762                         & 8.8611          & 0.2561                         & 3.6518          & 0.0243                         & 10.0324          & 0.0557                         & 4.3207                     & 0.0402                         & 8.1679                     & 0.0817                         \\
IDAM                   & 4.4153          & 0.1385                         & 8.6178          & 0.2756                         & 5.6727          & 0.0303                         & 11.5949          & 0.0629                         & 1.6348                     & 0.0230                         & 3.8151                     & 0.0491                         \\
RPM-Net                & 0.3267          & 0.0125                         & 0.6277          & 0.0246                         & 0.3885          & 0.0021                         & 0.7649           & 0.0042                         & 0.9164                     & 0.0146                         & 2.1291                     & 0.0303                         \\
STORM                  & 0.0311          & 0.0017                         & 0.1068          & 0.0057                         & 0.0304          & 0.0002                         & 0.0354           & 0.0001                         & 0.6624                     & 0.0147                         & 2.0765                     & 0.0512                         \\
RIENet                 & 0.0492          & 0.0023                         & 0.0897          & 0.0049                         & 0.0121          & 0.0001                         & 0.0299           & 0.0001                         & 0.8251                     & 0.0183                         & 1.8754                     & 0.0414                         \\ \hline
Ours                   & \textbf{0.0210} & \textbf{0.0013}                & \textbf{0.0664} & \textbf{0.0038}                & \textbf{0.0053} & \textbf{0.000034}              & \textbf{0.02304} & \textbf{0.000068}              & \textbf{0.5237}            & \textbf{0.0103}                & \textbf{1.5884}            & \textbf{0.0393}                \\ \hline
\end{tabular}
}
}
\caption{The registration results of different methods on the indoor scenes and outdoor dataset.}
\end{table*}

\begin{figure*}
	\centering
	\includegraphics[width=0.94\textwidth]{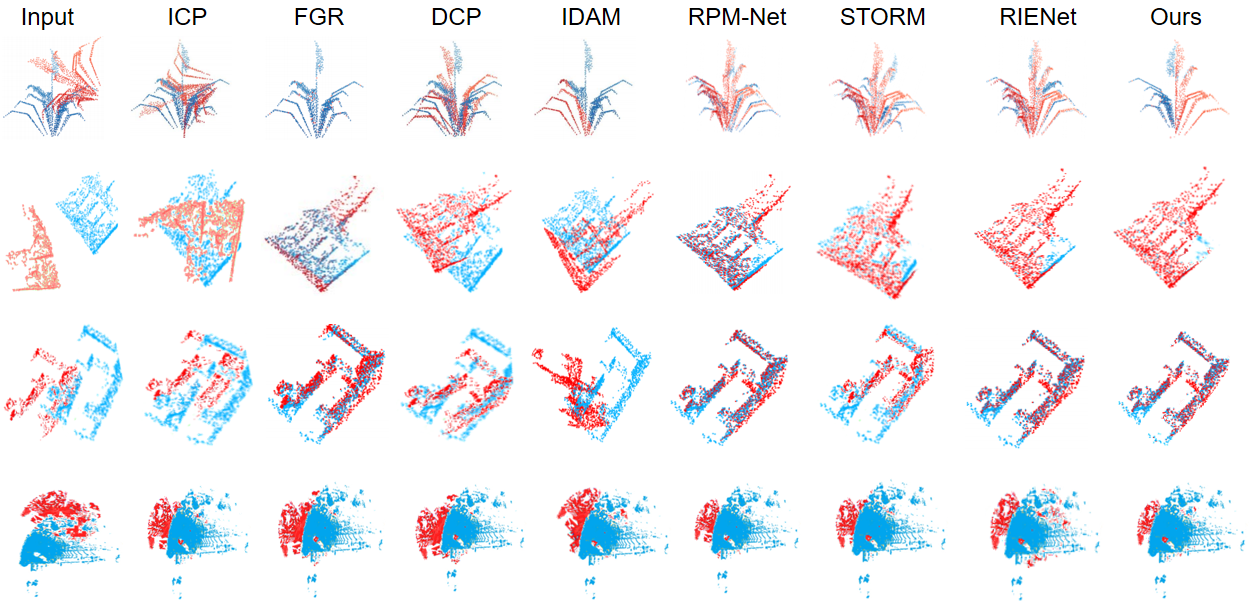}  
	\caption{Visualization of registration results on the ModelNet40, 7Scenes, ICL-NUIM and KITTI datasets.}
	\label{fig6}
	\vspace{-.1cm}
\end{figure*}

\begin{table}
\centering
\renewcommand\arraystretch{1.6}
\setlength{\tabcolsep}{1mm}{
\resizebox{0.99\columnwidth}{!}{
\begin{tabular}{ccccc|cccc}
\hline
\multirow{2}{*}{Model} & \multicolumn{4}{c|}{ModelNet40}                                     & \multicolumn{4}{c}{KITTI}                                         \\ \cline{2-9} 
                       & MAE(R)         & MAE(t)           & MIE(R)        & MIE(t)          & MAE(R)        & MAE(t)          & MIE(R)        & MIE(t)          \\ \hline
ICP                    & 11.67          & 0.16             & 23.04         & 0.33            & 15.37         & 12.65           & 39.67         & 30.10           \\
FGR                    & 4.74           & 0.062            & 9.53          & 0.128           & 17.23         & 1.3348          & 38.85         & 3.004           \\
DCP                    & 3.76           & 0.059            & 7.01          & 0.062           & 10.24         & 0.388           & 22.09         & 0.864           \\
IDAM                   & 7.02           & 0.110            & 14.86         & 0.233           & 18.09         & 0.8958          & 34.85         & 1.742           \\
RPM-Net                & 0.97           & 0.0064           & 3.12          & 0.0071          & 3.94          & 0.0976          & 7.26          & 0.305           \\
STORM                  & 0.134          & 0.00058          & 1.43          & 0.0017          & 1.27          & 0.0237          & 3.25          & 0.0867          \\
RIENet                 & 0.174          & 0.00062          & 0.88          & 0.0043          & 3.19          & 0.0664          & 5.15          & 0.177           \\ \hline
Ours                   & \textbf{0.057} & \textbf{0.00026} & \textbf{0.62} & \textbf{0.0007} & \textbf{1.05} & \textbf{0.0195} & \textbf{2.85} & \textbf{0.0574} \\ \hline
\end{tabular}
}}
\caption{The registration results of different methods on ModelNet40 and KITTI with overlap rate of 0.58.}
\end{table}

\section{Experiments}
	\subsection{Experimental Settings}
	\noindent \textbf{Datasets.} We evaluate our method on ModelNet40 ~\cite{wu20153d}, 7Scenes ~\cite{shotton2013scene}, ICL-NUIM ~\cite{choi2015robust} and KITTI odometry datasets ~\cite{geiger2012we}. The ModelNet40 consists of 12,311 meshed CAD models from 40 categories. We use 9,843 models for training and 2,468 models for testing. 7Scenes is a widely used benchmark registration dataset of indoor environment with 7 scenes including Chess, Fires, Heads, Office, Pumpkin, RedKitchen and Stairs.
	The dataset is divided into 296 samples for training and 57 for testing. 
	For another synthetic indoor scene ICL-NUIM, we first augment the dataset, then split the dataset into 1,278 samples for training and 200 samples for testing. And the KITTI odometry dataset consists of 11 sequences with ground truth pose, we use Sequence 00-05 for training, 06-07 for validation, and 08-10 for testing. We form the pairwise point clouds with the current frame and the 10th frame after it.
	
	\noindent \textbf{Compared methods and evaluation metrics.} We compare with traditional methods including ICP ~\cite{besl1992method}, FGR ~\cite{zhou2016fast}. Besides, we compare with deep methods, including supervised DCP~\cite{wang2019deep}, IDAM ~\cite{li2020iterative},RPM-Net ~\cite{yew2020rpm} and STORM  ~\cite{9705149}, and unsupervised RIENet ~\cite{2022Reliable}. 
	We evaluate the registration by the mean absolute errors (MAE) of $\mathbf{R}$ and $\mathbf{t}$ , which are anisotropic, and the mean isotropic errors (MIE) of $\mathbf{R}$ and $\mathbf{t}$ used in RIENet.
	
	\noindent \textbf{Implementation details.} Our model is implemented in Pytorch. We optimize the parameters with the ADAM optimizer. The initial learning rate is 0.001. For ModelNet40 and KITTI, we train the network for 50 epochs and multiply the learning rate by 0.7 at epoch 25. For indoor scenes, we multiply the learning rate by 0.7 at epochs 25, 50, 75 and train the network for 100 epochs.

\subsection{Comparison Evaluation}
\noindent \textbf{Analysis.}
We compared the registration performance of different methods on the ModelNet40 dataset under scenarios of ``Same categories'' , ``Unseen categories'' and ``Noise categories'', as shown in Table 1. Each source point cloud, denoted as P, in the training set, is transformed into Q using randomly generated transformation matrices. It can be observed that as an unsupervised learning approach, our method achieves the smallest errors among both traditional and deep learning methods.

We validated our approach on various real-world datasets, including ICL-NUIM and 7Scenes for indoor scenes, as well as KITTI for outdoor scenes.   Following the data processing in RIENet, our model shows promising performance in both indoor and outdoor scenarios, as shown in Table 2.    While the STORM method exhibits similarities with our approach in certain metrics, our method boasts nearly 1/5 of its runtime.  The visual outcomes across four distinct datasets are depicted in Figure 6.

\begin{table}
\centering
\renewcommand\arraystretch{1.4}
\setlength{\tabcolsep}{2mm}{
\resizebox{0.99\columnwidth}{!}{
\begin{tabular}{ccccc}
\hline
\multirow{2}{*}{Model} & \multicolumn{4}{c}{ModelNet40}                        \\ \cline{2-5} 
                        & MAE(R) & MAE(t) & MIE(R) & MIE(t) \\ \hline
OBMM+FGR                & 0.2714 & 0.0010 & 0.8236 & 0.0028 \\
OBMM+DCP               & 0.8842 & 0.0050 & 2.1410 & 0.0049 \\
OBMM+IDAM               & 0.0954 & 0.0008 & 0.1956 & 0.0009 \\
OBMM+RPM-Net           & 0.0050 & 0.00004 & 0.0221 & 0.00006 \\ \hline
\end{tabular}}}
\caption{The registration results of different methods combined with OBMM on ModelNet40.}
\vspace{-.5cm}
\end{table}

\noindent \textbf{Partial object registration.}
To validate the effectiveness of our method in partial-to-partial registration, we conducted experiments on the sparsely overlapped ModelNet40 and KITTI datasets. By employing a truncation approach, we generated point clouds with an approximate overlap rate of 0.58, where the point cloud overlap rate is defined as the ratio of point pairs in similar spatial positions to the total number of points. The experimental results, as presented in Table 3, demonstrate our method's capability to accurately detect point cloud overlap and generate partial correspondences. In comparison to other approaches, our method exhibits a notably pronounced advantage.

\noindent \textbf{Transferrability of OBMM.}
OBMM demonstrates impressive transferability, effectively improving the precision of various deep methods for partial-to-partial registration.  We integrated OBMM into FGR, DCP, IDAM, and RPM-Net as an overlap feature extractor for point clouds.  By predicting overlap biases, we introduced bias coefficients into the subsequent network to aid matrix transformation.  The results in Table 4 clearly illustrate the substantial enhancement in registration accuracy achieved through OBMM integration.  Remarkably, this improvement comes without significant additional runtime overhead.

\begin{table}
\centering
\renewcommand\arraystretch{1.5}
\setlength{\tabcolsep}{2mm}{
\resizebox{0.99\columnwidth}{!}{
\begin{tabular}{ccccccc}
\hline
NMM & \multicolumn{1}{l}{OS} & BP & MAE(R) & MAE(t) & MIE(R) & MIE(t) \\ \hline
    & \checkmark                      & \checkmark  & 0.0124 & 0.000035 & 0.098  & 0.000086 \\
\checkmark   &                        &    & 0.0058 & 0.000037 & 0.022  & 0.000103 \\
\checkmark   & \checkmark                      &    & 0.0043 & 0.000028 & 0.022  & 0.000052 \\
\checkmark   & \checkmark                      & \checkmark  & \textbf{0.0016} & \textbf{0.000014} & \textbf{0.019}  & \textbf{0.000031} \\ \hline
\end{tabular}}}
\caption{The results of different combination of key components on ModelNet40.}
\end{table}

\subsection{Ablation Study}
	\noindent \textbf {The effectiveness of key components.}
	We conducted an ablation study on the three key components of this model: Overlapping Sampling (OS), Bias Prediction (BP), and Neighbor Map Matching (NMM) modules. In the baseline configuration, we concatenated the point-wise features of the encoder and used them as input for the MLP, thereby replacing the OS module and eliminating the NMM module. The baseline model was trained using our unsupervised loss function. By incrementally adding these components, we obtained results for different categories on ModelNet40 dataset. The outcomes are presented in Table 5. Notably, the early OS module exhibits significant improvements over the baseline. Furthermore, the neighbor map matching module contributes to achieving more precise partial data registration results.

	\noindent \textbf{Loss functions.}
	In our experiments, we train our model with the combination of the global alignment loss $\mathcal{L}_{g}$, the neighborhood agreement loss $\mathcal{L}_{n}$, and the spatial consistency loss $\mathcal{L}_{s}$. In Table 6, we train our model using the different loss functions and present the results on KITTI. One can see that with the neighborhood consensus loss and spatial consistency loss based on the selected module, we can obtain high registration precision.

\begin{table}
\centering
\renewcommand\arraystretch{1.5}
\setlength{\tabcolsep}{2mm}{
\resizebox{0.99\columnwidth}{!}{
\begin{tabular}{lcccc}
\hline
\multicolumn{1}{c}{Loss} & MAE(R) & MAE(t) & MIE(R)  & MIE(t) \\ \hline
 $\mathcal{L}_{g}$                      & 2.0456 & 0.9888 & 74.2689 & 5.6725 \\
$\mathcal{L}_{g}+\mathcal{L}_{s}$                  & 4.8653 & 0.2195 & 14.8652 & 1.1569 \\
$\mathcal{L}_{g}+\mathcal{L}_{s}+\mathcal{L}_{n}$              & \textbf{0.5237} & \textbf{0.0103} & \textbf{1.5884}  &\textbf{ 0.0393} \\ \hline
\end{tabular}}}
\caption{The results of different combination of loss functions on KITTI.}
\end{table}
\vspace{-.2cm}

\section{Conclusion}
In this paper, we propose a plug-and-play Overlap Bias Matching Module (OBMM) and then integrate it with the neighbor map matching module to construct an unsupervised network for partial point cloud registration. The OBMM comprises two components: a differentiable overlap sampling module and a bias prediction module. The proposed OBMM has good generalization and can be easily integrated into most registration networks. By integrating OBMM and the neighbor map matching module, we have devised an unsupervised point cloud registration network. This network can maintain efficacy even in pair-wise registration scenarios with low overlap ratios.     Experimental results on extensive datasets demonstrate that our method achieves state-of-the-art registration performance.

\bigskip
\noindent 

\bibliography{aaai24}

\begin{thebibliography}{55}
\providecommand{\natexlab}[1]{#1}

\bibitem[{Agarwal et~al.(2011)Agarwal, Furukawa, Snavely, Simon, Curless,
  Seitz, and Szeliski}]{agarwal2011building}
Agarwal, S.; Furukawa, Y.; Snavely, N.; Simon, I.; Curless, B.; Seitz, S.~M.;
  and Szeliski, R. 2011.
\newblock Building rome in a day.
\newblock \emph{Communications of the ACM}, 54(10): 105--112.

\bibitem[{Ali et~al.(2021)Ali, Kahraman, Reis, and Stricker}]{ali2021rpsrnet}
Ali, S.~A.; Kahraman, K.; Reis, G.; and Stricker, D. 2021.
\newblock RPSRNet: End-to-End Trainable Rigid Point Set Registration Network
  using Barnes-Hut 2D-Tree Representation.
\newblock In \emph{CVPR}.

\bibitem[{Aoki et~al.(2019)Aoki, Goforth, Srivatsan, and
  Lucey}]{aoki2019pointnetlk}
Aoki, Y.; Goforth, H.; Srivatsan, R.~A.; and Lucey, S. 2019.
\newblock PointNetLK: Robust \& efficient point cloud registration using
  pointnet.
\newblock In \emph{CVPR}.

\bibitem[{Bai et~al.(2021)Bai, Luo, Zhou, Chen, Li, Hu, Fu, and
  Tai}]{bai2021pointdsc}
Bai, X.; Luo, Z.; Zhou, L.; Chen, H.; Li, L.; Hu, Z.; Fu, H.; and Tai, C.-L.
  2021.
\newblock PointDSC: Robust Point Cloud Registration using Deep Spatial
  Consistency.
\newblock In \emph{CVPR}.

\bibitem[{Besl and McKay(1992)}]{besl1992method}
Besl, P.~J.; and McKay, N.~D. 1992.
\newblock Method for registration of 3-{D} shapes.
\newblock In \emph{SPIE}.

\bibitem[{Bouaziz, Tagliasacchi, and Pauly(2013)}]{bouaziz2013sparse}
Bouaziz, S.; Tagliasacchi, A.; and Pauly, M. 2013.
\newblock Sparse iterative closest point.
\newblock In \emph{Computer graphics forum}, volume~32, 113--123.

\bibitem[{Chen, Hung, and Cheng(1999)}]{chen1999ransac}
Chen, C.-S.; Hung, Y.-P.; and Cheng, J.-B. 1999.
\newblock RANSAC-based DARCES: A new approach to fast automatic registration of
  partially overlapping range images.
\newblock \emph{IEEE Transactions on Pattern Analysis and Machine
  Intelligence}, 21(11): 1229--1234.

\bibitem[{Choi, Zhou, and Koltun(2015)}]{choi2015robust}
Choi, S.; Zhou, Q.-Y.; and Koltun, V. 2015.
\newblock Robust reconstruction of indoor scenes.
\newblock In \emph{CVPR}.

\bibitem[{Choy, Dong, and Koltun(2020)}]{choy2020deep}
Choy, C.; Dong, W.; and Koltun, V. 2020.
\newblock Deep Global Registration.
\newblock In \emph{CVPR}.

\bibitem[{Deng, Birdal, and Ilic(2019)}]{deng20193d}
Deng, H.; Birdal, T.; and Ilic, S. 2019.
\newblock 3D local features for direct pairwise registration.
\newblock In \emph{CVPR}.

\bibitem[{Deschaud(2018)}]{deschaud2018imls}
Deschaud, J.-E. 2018.
\newblock IMLS-SLAM: scan-to-model matching based on 3D data.
\newblock In \emph{ICRA}.

\bibitem[{El~Banani, Gao, and Johnson(2021)}]{el2021unsupervisedr}
El~Banani, M.; Gao, L.; and Johnson, J. 2021.
\newblock Unsupervisedr\&r: Unsupervised point cloud registration via
  differentiable rendering.
\newblock In \emph{CVPR}.

\bibitem[{Feng et~al.(2021)Feng, Zhang, Cai, Xu, Hou, and
  Bao}]{feng2021recurrent}
Feng, W.; Zhang, J.; Cai, H.; Xu, H.; Hou, J.; and Bao, H. 2021.
\newblock Recurrent Multi-view Alignment Network for Unsupervised Surface
  Registration.
\newblock In \emph{CVPR}.

\bibitem[{Fischer et~al.(2021)Fischer, Simon, Olsner, Milz, Gross, and
  Mader}]{fischer2021stickypillars}
Fischer, K.; Simon, M.; Olsner, F.; Milz, S.; Gross, H.-M.; and Mader, P. 2021.
\newblock Stickypillars: Robust and efficient feature matching on point clouds
  using graph neural networks.
\newblock In \emph{CVPR}.

\bibitem[{Fischler and Bolles(1981)}]{fischler1981random}
Fischler, M.~A.; and Bolles, R.~C. 1981.
\newblock Random sample consensus: a paradigm for model fitting with
  applications to image analysis and automated cartography.
\newblock \emph{Communications of the ACM}, 24(6): 381--395.

\bibitem[{Fu et~al.(2021)Fu, Liu, Luo, and Wang}]{fu2021robust}
Fu, K.; Liu, S.; Luo, X.; and Wang, M. 2021.
\newblock Robust Point Cloud Registration Framework Based on Deep Graph
  Matching.
\newblock In \emph{CVPR}.

\bibitem[{Geiger, Lenz, and Urtasun(2012)}]{geiger2012we}
Geiger, A.; Lenz, P.; and Urtasun, R. 2012.
\newblock Are we ready for autonomous driving? the kitti vision benchmark
  suite.
\newblock In \emph{CVPR}.

\bibitem[{Gojcic et~al.(2020)Gojcic, Zhou, Wegner, Guibas, and
  Birdal}]{gojcic2020learning}
Gojcic, Z.; Zhou, C.; Wegner, J.~D.; Guibas, L.~J.; and Birdal, T. 2020.
\newblock Learning multiview 3D point cloud registration.
\newblock In \emph{CVPR}.

\bibitem[{Groueix et~al.(2019)Groueix, Fisher, Kim, Russell, and
  Aubry}]{groueix2019unsupervised}
Groueix, T.; Fisher, M.; Kim, V.~G.; Russell, B.~C.; and Aubry, M. 2019.
\newblock Unsupervised cycle-consistent deformation for shape matching.
\newblock In \emph{Computer Graphics Forum}, volume~38, 123--133.

\bibitem[{Huang et~al.(2021)Huang, Gojcic, Usvyatsov, Wieser, and
  Schindler}]{huang2021predator}
Huang, S.; Gojcic, Z.; Usvyatsov, M.; Wieser, A.; and Schindler, K. 2021.
\newblock PREDATOR: Registration of 3D Point Clouds with Low Overlap.
\newblock In \emph{CVPR}.

\bibitem[{Huang, Mei, and Zhang(2020)}]{huang2020feature}
Huang, X.; Mei, G.; and Zhang, J. 2020.
\newblock Feature-metric Registration: A Fast Semi-supervised Approach for
  Robust Point Cloud Registration without Correspondences.
\newblock In \emph{CVPR}.

\bibitem[{Jiang et~al.(2021)Jiang, Shen, Li, Qian, Xie, and
  Yang}]{jiang2021CEMNet}
Jiang, H.; Shen, Y.; Li, J.; Qian, J.; Xie, J.; and Yang, J. 2021.
\newblock Sampling Network Guided Cross-Entropy Method for Unsupervised Point
  Cloud Registration.
\newblock In \emph{ICCV}.

\bibitem[{Kadam et~al.(2020)Kadam, Zhang, Liu, and Kuo}]{kadam2020unsupervised}
Kadam, P.; Zhang, M.; Liu, S.; and Kuo, C.-C.~J. 2020.
\newblock Unsupervised Point Cloud Registration via Salient Points Analysis
  (SPA).
\newblock In \emph{VCIP}.

\bibitem[{Le et~al.(2019)Le, Do, Hoang, and Cheung}]{le2019sdrsac}
Le, H.~M.; Do, T.-T.; Hoang, T.; and Cheung, N.-M. 2019.
\newblock Sdrsac: Semidefinite-based randomized approach for robust point cloud
  registration without correspondences.
\newblock In \emph{CVPR}.

\bibitem[{Li et~al.(2020)Li, Zhang, Xu, Zhou, and Zhang}]{li2020iterative}
Li, J.; Zhang, C.; Xu, Z.; Zhou, H.; and Zhang, C. 2020.
\newblock Iterative distance-aware similarity matrix convolution with
  mutual-supervised point elimination for efficient point cloud registration.
\newblock In \emph{Computer Vision--ECCV 2020: 16th European Conference,
  Glasgow, UK, August 23--28, 2020, Proceedings, Part XXIV 16}, 378--394.
  Springer.

\bibitem[{Li, Pontes, and Lucey(2021)}]{li2021pointnetlk}
Li, X.; Pontes, J.~K.; and Lucey, S. 2021.
\newblock PointNetLK Revisited.
\newblock In \emph{CVPR}.

\bibitem[{Li, Wang, and Fang(2019)}]{li2019pc}
Li, X.; Wang, L.; and Fang, Y. 2019.
\newblock Pc-net: Unsupervised point correspondence learning with neural
  networks.
\newblock In \emph{3DV}.

\bibitem[{Lu et~al.(2019)Lu, Wan, Zhou, Fu, Yuan, and Song}]{lu2019deepvcp}
Lu, W.; Wan, G.; Zhou, Y.; Fu, X.; Yuan, P.; and Song, S. 2019.
\newblock DeepVCP: An End-to-End Deep Neural Network for Point Cloud
  Registration.
\newblock In \emph{ICCV}.

\bibitem[{Mellado, Aiger, and Mitra(2014)}]{mellado2014super}
Mellado, N.; Aiger, D.; and Mitra, N.~J. 2014.
\newblock Super 4pcs fast global pointcloud registration via smart indexing.
\newblock In \emph{Computer graphics forum}, volume~33, 205--215.

\bibitem[{Pais et~al.(2020)Pais, Ramalingam, Govindu, Nascimento, Chellappa,
  and Miraldo}]{pais20203dregnet}
Pais, G.~D.; Ramalingam, S.; Govindu, V.~M.; Nascimento, J.~C.; Chellappa, R.;
  and Miraldo, P. 2020.
\newblock 3DRegNet: A deep neural network for 3D point registration.
\newblock In \emph{CVPR}.

\bibitem[{Phillips, Liu, and Tomasi(2007)}]{phillips2007outlier}
Phillips, J.~M.; Liu, R.; and Tomasi, C. 2007.
\newblock Outlier robust ICP for minimizing fractional RMSD.
\newblock In \emph{3DIM}.

\bibitem[{Qi et~al.(2017{\natexlab{a}})Qi, Su, Mo, and Guibas}]{qi2017pointnet}
Qi, C.~R.; Su, H.; Mo, K.; and Guibas, L.~J. 2017{\natexlab{a}}.
\newblock PointNet: Deep learning on point sets for 3D classification and
  segmentation.
\newblock In \emph{CVPR}.

\bibitem[{Qi et~al.(2017{\natexlab{b}})Qi, Yi, Su, and
  Guibas}]{qi2017pointnet++}
Qi, C.~R.; Yi, L.; Su, H.; and Guibas, L.~J. 2017{\natexlab{b}}.
\newblock PointNet++: Deep hierarchical feature learning on point sets in a
  metric space.
\newblock In \emph{NIPS}.

\bibitem[{Rusu, Blodow, and Beetz(2009)}]{rusu2009fast}
Rusu, R.~B.; Blodow, N.; and Beetz, M. 2009.
\newblock Fast point feature histograms (FPFH) for 3D registration.
\newblock In \emph{ICRA}.

\bibitem[{Rusu et~al.(2008)Rusu, Blodow, Marton, and Beetz}]{rusu2008aligning}
Rusu, R.~B.; Blodow, N.; Marton, Z.~C.; and Beetz, M. 2008.
\newblock Aligning point cloud views using persistent feature histograms.
\newblock In \emph{IROS}.

\bibitem[{Sarode et~al.(2020)Sarode, Dhagat, Srivatsan, Zevallos, Lucey, and
  Choset}]{sarode2020masknet}
Sarode, V.; Dhagat, A.; Srivatsan, R.~A.; Zevallos, N.; Lucey, S.; and Choset,
  H. 2020.
\newblock MaskNet: A Fully-Convolutional Network to Estimate Inlier Points.
\newblock \emph{arXiv preprint arXiv:2010.09185}.

\bibitem[{Sarode et~al.(2019)Sarode, Li, Goforth, Aoki, Srivatsan, Lucey, and
  Choset}]{sarode2019pcrnet}
Sarode, V.; Li, X.; Goforth, H.; Aoki, Y.; Srivatsan, R.~A.; Lucey, S.; and
  Choset, H. 2019.
\newblock PCRNet: Point cloud registration network using PointNet encoding.
\newblock \emph{arXiv preprint arXiv:1908.07906}.

\bibitem[{Schonberger and Frahm(2016)}]{schonberger2016structure}
Schonberger, J.~L.; and Frahm, J.-M. 2016.
\newblock Structure-from-motion revisited.
\newblock In \emph{CVPR}.

\bibitem[{Segal, Haehnel, and Thrun(2009)}]{segal2009generalized}
Segal, A.; Haehnel, D.; and Thrun, S. 2009.
\newblock Generalized-icp.
\newblock In \emph{RSS}.

\bibitem[{Shen et~al.(2022)Shen, Hui, Jiang, Xie, and Yang}]{2022Reliable}
Shen, Y.; Hui, L.; Jiang, H.; Xie, J.; and Yang, J. 2022.
\newblock Reliable Inlier Evaluation for Unsupervised Point Cloud Registration.
\newblock \emph{arXiv e-prints}.

\bibitem[{Shotton et~al.(2013)Shotton, Glocker, Zach, Izadi, Criminisi, and
  Fitzgibbon}]{shotton2013scene}
Shotton, J.; Glocker, B.; Zach, C.; Izadi, S.; Criminisi, A.; and Fitzgibbon,
  A. 2013.
\newblock Scene coordinate regression forests for camera relocalization in
  RGB-D images.
\newblock In \emph{CVPR}.

\bibitem[{Tombari, Salti, and Di~Stefano(2010)}]{tombari2010unique}
Tombari, F.; Salti, S.; and Di~Stefano, L. 2010.
\newblock Unique shape context for 3D data description.
\newblock In \emph{ACM workshop}.

\bibitem[{Wang, Li, and Fang(2020)}]{wang2020unsupervised}
Wang, L.; Li, X.; and Fang, Y. 2020.
\newblock Unsupervised Learning of 3D Point Set Registration.
\newblock \emph{arXiv preprint arXiv:2006.06200}.

\bibitem[{Wang and Solomon(2019{\natexlab{a}})}]{wang2019deep}
Wang, Y.; and Solomon, J.~M. 2019{\natexlab{a}}.
\newblock Deep closest point: Learning representations for point cloud
  registration.
\newblock In \emph{Proceedings of the IEEE/CVF international conference on
  computer vision}, 3523--3532.

\bibitem[{Wang and Solomon(2019{\natexlab{b}})}]{wang2019prnet}
Wang, Y.; and Solomon, J.~M. 2019{\natexlab{b}}.
\newblock PRNet: Self-supervised learning for partial-to-partial registration.
\newblock In \emph{NIPS}.

\bibitem[{Wang et~al.(2023)Wang, Yan, Feng, Du, Dai, and Gao}]{9705149}
Wang, Y.; Yan, C.; Feng, Y.; Du, S.; Dai, Q.; and Gao, Y. 2023.
\newblock STORM: Structure-Based Overlap Matching for Partial Point Cloud
  Registration.
\newblock \emph{IEEE Transactions on Pattern Analysis and Machine
  Intelligence}, 45(1): 1135--1149.

\bibitem[{Wong et~al.(2017)Wong, Kee, Le, Wagner, Mariottini, Schneider,
  Hamilton, Chipalkatty, Hebert, Johnson et~al.}]{wong2017segicp}
Wong, J.~M.; Kee, V.; Le, T.; Wagner, S.; Mariottini, G.-L.; Schneider, A.;
  Hamilton, L.; Chipalkatty, R.; Hebert, M.; Johnson, D.~M.; et~al. 2017.
\newblock SegICP: Integrated deep semantic segmentation and pose estimation.
\newblock In \emph{IROS}.

\bibitem[{Wu et~al.(2015)Wu, Song, Khosla, Yu, Zhang, Tang, and
  Xiao}]{wu20153d}
Wu, Z.; Song, S.; Khosla, A.; Yu, F.; Zhang, L.; Tang, X.; and Xiao, J. 2015.
\newblock 3D ShapeNets: A Deep Representation for Volumetric Shapes.
\newblock In \emph{CVPR}.

\bibitem[{Yang, Li, and Jia(2013)}]{yang2013go}
Yang, J.; Li, H.; and Jia, Y. 2013.
\newblock Go-ICP: Solving 3D Registration Efficiently and Globally Optimally.
\newblock In \emph{ICCV}.

\bibitem[{Yang et~al.(2020)Yang, Liu, Cui, Chen, and Wang}]{yang2020mapping}
Yang, L.; Liu, W.; Cui, Z.; Chen, N.; and Wang, W. 2020.
\newblock Mapping in a cycle: Sinkhorn regularized unsupervised learning for
  point cloud shapes.
\newblock In \emph{ECCV}.

\bibitem[{Yang et~al.(2017)Yang, Feng, Shen, and Tian}]{yang2017foldingnet}
Yang, Y.; Feng, C.; Shen, Y.; and Tian, D. 2017.
\newblock Foldingnet: Interpretable unsupervised learning on 3d point clouds.
\newblock \emph{arXiv preprint arXiv:1712.07262}.

\bibitem[{Yew and Lee(2020)}]{yew2020rpm}
Yew, Z.~J.; and Lee, G.~H. 2020.
\newblock RPM-Net: Robust Point Matching using Learned Features.
\newblock In \emph{CVPR}.

\bibitem[{Yuan et~al.(2020)Yuan, Eckart, Kim, Jampani, Fox, and
  Kautz}]{yuan2020deepgmr}
Yuan, W.; Eckart, B.; Kim, K.; Jampani, V.; Fox, D.; and Kautz, J. 2020.
\newblock Deepgmr: Learning latent gaussian mixture models for registration.
\newblock In \emph{ECCV}.

\bibitem[{Zhang and Singh(2014)}]{zhang2014loam}
Zhang, J.; and Singh, S. 2014.
\newblock LOAM: Lidar Odometry and Mapping in Real-time.
\newblock In \emph{RSS}.

\bibitem[{Zhou, Park, and Koltun(2016)}]{zhou2016fast}
Zhou, Q.-Y.; Park, J.; and Koltun, V. 2016.
\newblock Fast global registration.
\newblock In \emph{ECCV}.

\end{thebibliography}

\end{document}